\documentclass[letterpaper]{article} 
\usepackage{aaai2026}  
\usepackage{times}  
\usepackage{helvet}  
\usepackage{courier}  
\usepackage[hyphens]{url}  
\usepackage{graphicx} 
\usepackage{threeparttable}
\usepackage{natbib}  
\usepackage{caption} 
\usepackage{algorithm}
\usepackage{algorithmic}
\usepackage{xcolor}
\usepackage{booktabs}   
\usepackage{siunitx}    
\usepackage{caption}    
\usepackage{adjustbox}
\usepackage{newfloat}
\usepackage{listings}
\usepackage{xcolor}
\usepackage[T1]{fontenc}
\usepackage{inconsolata}  
\usepackage{ragged2e}
\usepackage{float}
\usepackage{booktabs} 
\usepackage{graphicx}

\usepackage{microtype}

\usepackage[most]{tcolorbox}     
\usepackage{enumitem}

\tcbuselibrary{listings,skins}

\usepackage[T1]{fontenc}
\usepackage[utf8]{inputenc}   
\usepackage{inconsolata}      

\usepackage{lipsum}  

\DeclareCaptionStyle{ruled}
{labelfont=normalfont,labelsep=colon,strut=off} 
\floatstyle{ruled}
\newfloat{listing}{tb}{lst}{}
\floatname{listing}{Listing}
\providecommand{\ensuretext}[1]{\ifmmode\text{#1}\else#1\fi}

\newif\ifcomments
\commentstrue   

\ifcomments
  \newcommand{\draftcomment}[3]{\ensuretext{\textcolor{#2}{[\textsc{#1} #3]}}}
\else
  \newcommand{\draftcomment}[3]{}
\fi

\title{Dharma, Data and Deception: An LLM-Powered Rhetorical Analysis of Cow-Urine Health Claims on YouTube}

\author{
 Sheza Munir \textsuperscript{\rm 1}, Ratna Kandala\textsuperscript{\rm 2}, Anamta Khan\textsuperscript{\rm 1}, Deepti\textsuperscript{\rm 3}, Joyojeet Pal\textsuperscript{\rm 1}
}
\affiliations{
    \textsuperscript{\rm 1} University of Michigan, \textsuperscript{\rm 2}University of Kansas, \textsuperscript{\rm 3}IIT Jodhpur
}

\begin{document}
\maketitle

\begin{abstract}
Health misinformation remains one of the most pressing challenges on social media, particularly when cultural traditions intersect with scientific-sounding claims. These dynamics are not only global but also deeply local, manifesting in culturally specific controversies that require careful analysis. Motivated by this, we examine 100 YouTube transcripts that promote or debunk cow urine (gomutra) as a health remedy, focusing on rhetorical strategies such as appeals to authority, efficacy appeals, and conspiracy framing. We employ large language models (LLMs) including GPT-4, GPT-4o, GPT-4.1, GPT-5, Gemini 2.5 Pro, and Mistral Medium 3 to annotate transcripts using a 14-category taxonomy of persuasive tactics. Our analysis reveals that promoters predominantly rely on efficacy appeals and social proof, while debunkers emphasize authority and rebuttal.
Human evaluation of a subset of annotations yielded 90.1\% inter-annotator agreement, confirming the reliability of our taxonomy and validation process. This work advances computational methods for misinformation analysis and demonstrates how LLMs can support large-scale studies of cultural discourse online.

\end{abstract}

\section{Introduction}

Health misinformation on social media erodes confidence in medicine and public health \cite{vosoughi2018spread,bennett2019disinformation}. On high-reach platforms like YouTube, claims about prevention and cures diffuse through creator networks and recommendation systems, where persuasive narratives often mix scientific-sounding language with cultural authority, personal testimony, and conspiratorial framing \cite{cinelli2020infodemic,germani2021anti}. In such environments, audiences encounter messages that are not only factually disputable but also rhetorically compelling: crafted to be memorable, identity-relevant, and easy to share.

This paper examines a culturally specific case: the Indian cow-urine (gomutra) discourse on YouTube. Gomutra has been promoted in some Indian contexts as a preventive or curative substance, while others criticize such claims as pseudoscientific. The debate sits at the intersection of religious tradition, national identity, and biomedical authority, making it a useful lens on how culture and credibility are negotiated in public health talk. The discourse is especially salient in India today, given the centrality of the bovine in certain Hindu traditions and its entanglement with contemporary politics \cite{jakobsen2023bovine}.

Prior work on conspiracy thinking, identity, and motivated reasoning suggests that appeals to group belonging and perceptions that elites are suppressing the truth shape belief formation \cite{douglas2019conspiracy}, yet large-scale, tactic-level descriptions of \emph{how} these persuasive resources are deployed within specific health debates on YouTube remain scarce. Building on persuasion theory and health communication, this study argues that measuring \emph{how} a claim persuades (its strategic tactics) is as important as measuring \emph{what} stance it takes. Classic models highlight central and peripheral routes to attitude change \cite{petty1986elaboration}, perceived threat and efficacy in fear appeals \cite{witte2000meta}, perceived benefits and barriers in the Health Belief Model \cite{rosenstock1974health}, and influence principles such as authority and social proof \cite{cialdini2007influence}. Critical discourse analysis explains how appeals to authority and tradition construct legitimacy in contested settings \cite{van1993elite}. We operationalize these insights by defining a 14-category taxonomy of persuasive tactics, ranging from authority, tradition/religion, and efficacy appeals to scientific jargon, conspiracy framing, and social proof, that can be systematically observed in text.

A persistent challenge is scale: manual discourse annotation is time-consuming and requires domain and cultural knowledge. LLMs can accelerate annotation but raise concerns about bias and cultural sensitivity \cite{gilardi2023chatgpt,santurkar2023whose}. This study adopts an LLM-assisted pipeline that inserts inline tactic tags into full transcripts, preserving context for corpus-level quantification, and then validates a subset with human annotators to characterize precision-oriented performance and to locate culturally sensitive boundaries. The analysis foregrounds cultural specificity: in this discourse, religious or community figures can simultaneously instantiate “authority” and “tradition,” code-mixing (e.g., Hindi–English) shifts surface cues, and ritual calendars shape when social proof is most salient; factors that bear directly on detection quality and interpretation.

This paper studies three questions:
\begin{enumerate}
  \item \textbf{RQ1:} How do persuasive tactics vary between promoting and debunking content in the gomutra discourse?
  \item \textbf{RQ2:} How do LLMs differ in sensitivity to culturally loaded tactics (e.g., tradition/religion versus scientific authority), and how does this shape their overall tagging profiles?
  \item \textbf{RQ3:} How well do LLM-generated annotations of persuasive tactics align with human judgments, and what does this reveal about the precision and limits of automated rhetorical analysis?
\end{enumerate}

To answer these questions, we analyze a corpus of 100 YouTube videos using a standardized prompt that returns full transcripts with inline tactic labels. We then compare the distributions and co-occurrences of persuasive tactics across stances and model families including GPT-4 \cite{openai_gpt4_2023}, GPT-4.1 \cite{openai2025gpt41}, GPT-4o \cite{openai_gpt4o_2024}, GPT-5 \cite{openai_gpt5_2025}, Gemini 2.5 Pro \cite{comanici2025gemini2.5}, and Mistral Medium 3 \cite{mistral2025medium3}, and evaluate a subset of annotations with human validators, focusing on agreement as a proxy for precision.

\subsection{Contributions:}
\begin{itemize}
  \item \textbf{A rich health misinformation corpus:} A curated corpus of 100 YouTube video transcripts on the gomutra discourse (with stance labels and video metadata), enabling reproducible, culture-specific analysis of health misinformation.
  \item \textbf{Taxonomy \& LLM analysis:} We introduce a 14-cue taxonomy of persuasive tactics in health-related misinformation and implement a scalable, auditable LLM-assisted annotation pipeline. We apply this pipeline and conduct a comparative study of persuasive tactic distributions across six state-of-the-art (SOTA) models (GPT-4, GPT-4.1, GPT-4o, GPT-5, Gemini~2.5~Pro, Mistral~Medium~3). We report stance-wise distributions, co-occurrences, and model divergence, identifying systematic differences in how models represent persuasion.
  \item \textbf{Human Evaluation:} We validate model outputs through a structured human evaluation that assesses both the reliability of the taxonomy and the precision of model-generated cues. This framework establishes a benchmark for evaluating automated persuasion analysis and underscores the value of combining LLM-scale annotation with human judgment.
\end{itemize}

\section{Related Work}

Research on misinformation has shown that false or misleading claims thrive not only because of their content, but also because of the persuasive forms in which they are framed. Studies of online health misinformation highlight how narratives that mix scientific jargon with cultural or religious authority can be especially compelling \cite{germani2021anti, chou2018addressing, borges2022infodemic}. 

A wide body of persuasion theory provides foundations for analyzing such rhetorical strategies. Classical Rhetoric emphasizes appeals to ethos, pathos, and logos as core persuasive modes, while Speech Act Theory highlights how utterances themselves function as actions that can assert, command, or persuade. The Elaboration Likelihood Model distinguishes between central and peripheral routes of persuasion, underscoring how heuristic cues such as social proof can drive belief formation when audiences are not motivated to engage deeply \cite{petty1986elaboration}. The Extended Parallel Process Model identifies the balance of threat and efficacy as key determinants of persuasive success in fear appeals \cite{witte2000meta}. Critical Discourse Analysis (CDA) situates persuasion within ideology and power, showing how authority, tradition, and in-group/out-group distinctions function to legitimize or marginalize claims \cite{waugh2015cda}. 

Critical discourse and political communication research further underscores the role of cultural identity and conspiracy framing in shaping belief formation \cite{douglas2019conspiracy, jakobsen2023bovine}. In the Indian context, cows carry both religious and political symbolism, making claims about gomutra a particularly salient case where biomedical credibility intersects with devotional authority. Work on information disorder has argued that counter-messaging cannot rely on factual correction alone but must also address rhetorical and emotional appeals \cite{chan2017debunking, wardle2017disorder}. 

Computational approaches to rhetorical analysis have evolved from small-scale, supervised pipelines to large-scale annotation frameworks. Early work in argumentation mining and persuasion detection emphasized handcrafted features and limited datasets \cite{tan2016winning, lippi2016argument, card2015mediaframes}. More recent studies demonstrate that large language models can outperform traditional methods and even crowdworkers on text annotation tasks \cite{gilardi2023chatgpt}, though cultural sensitivity and reliability remain open concerns \cite{santurkar2023whose}. Prior work also shows that LLMs perform best on explicit surface-level cues, while implicit or discourse-structured tactics often require human oversight \cite{kreps2022democracy, zhao2023llms}. 

Finally, platform-level research shows that the diffusion of misinformation is driven less by veracity than by resonance with identity and affect, amplified by algorithmic curation and echo chambers \cite{vosoughi2018spread, cinelli2020infodemic, garimella2017political, bennett2019disinformation}. These insights suggest that analyzing rhetorical tactics at scale, while accounting for cultural specificity, is essential for understanding how health misinformation circulates online. 


\begin{figure}[t]
  \centering
  \includegraphics[width=.7\linewidth]{"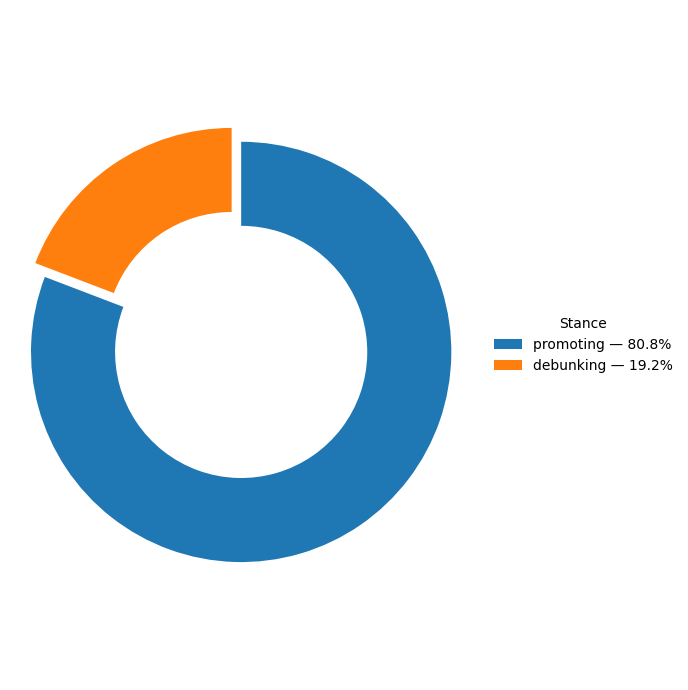"}
  \caption{Stance distribution of the dataset}
  \label{fig:stance-dist}
\end{figure}
\section{Methodology}

\subsection{Dataset}
To analyze discourse surrounding gomutra on social media, we gathered a corpus of YouTube videos. Data was retrieved using the YouTube Data API v3, accessed programmatically via Postman with an authenticated personal API key. A comprehensive search strategy was implemented to capture relevance and linguistic variation, employing combinations of keywords and hashtags including \textbf{gomutra}, \textbf{cow urine}, \textbf{cowurine}, \textbf{gaumutra}, \textbf{gaumutra health}, \textbf{gomutra health},  \textbf{\#gomutra}, \textbf{\#gomata} and \textbf{\#cowurine}. For each query, up to the API’s per-request maximum of 50 results were collected. The initial retrieval returned videos in multiple languages (e.g., English, Hindi, Urdu, Kannada, Tamil, and Guchjarati). We then applied a two-step curation process. First, to ensure topic relevance, we manually reviewed titles, descriptions, and audio, retaining only videos with primary spoken languages of English, Hindi, or Urdu. Second, we removed duplicates and derivative content to preserve dataset integrity. This multi-stage curation yielded a final set of 100 unique gomutra-related videos, with a total duration of 7 hours, 9 minutes, and 26 seconds.

After collecting the dataset, we extracted audio from all 100 YouTube videos and transcribed it into text for content analysis. Transcription was performed using OpenAI’s Whisper-Large model, selected for its state-of-the-art accuracy and strong performance across multiple languages \cite{radford2022robust}, which was essential given that the dataset includes videos in Hindi, English, and Urdu. To verify transcription quality, we conducted a human evaluation on a sampled subset. We computed word error rate (WER) over 6,098 transcribed words and observed an average WER of 6.5\%, which is lower than the 12.8\% average reported across multiple datasets by \cite{radford2022robust}, indicating that our transcripts are accurate and have a low error rate. To ensure data compatibility for analyses, all non-English transcripts were translated into English using GPT-4 via the OpenAI API. The model was run in a zero-shot configuration with a temperature of 0.7 to encourage more consistent and deterministic outputs. The model was specifically instructed to act as a culturally aware translator, preserving the source text’s meaning, tone, and cultural nuances while avoiding summarization or paraphrasing and leaving any existing English text unchanged.

Each video was manually annotated for its stance toward gomutra (promoting or debunking) in order to analyze persuasive techniques by stance. As shown in Figure~\ref{fig:stance-dist}, the distribution is highly skewed, with 80.8\% of videos promoting and 19.2\% debunking. Table~\ref{tab:dataset-stats} summarizes audience and engagement statistics. On average, videos are 4.5 minutes long ($sd = 5.8$, median = 2.12) and contain 456 words per transcript ($sd = 521$, median = 263). Engagement is also skewed: the mean number of likes is 17,086 ($sd = 64,145$, median = 1,585), mean views are 564,111 ($sd = 1,824,114$, median = 39,320), and mean comments are 1,022 ($sd = 6,098$, median = 64). Channels posting gomutra-related videos tend to have large followings, averaging 4.14M subscribers ($sd = 11.65M$, median = 533,000). The gaps between means, medians, and standard deviations highlight the variability and skewed distribution of the dataset.

\sisetup{group-separator={,}, group-minimum-digits=4}
\captionsetup[table]{justification=centering, singlelinecheck=false}

\begin{table}[t]

\begin{adjustbox}{max width=\linewidth}
\begin{tabular}{l r r r}
\toprule
\textbf{Statistic} & \multicolumn{1}{c}{\textbf{Mean}}
                   & \multicolumn{1}{c}{\textbf{Std.\ Dev.\ (sd)}}
                   & \multicolumn{1}{c}{\textbf{Median}} \\
\midrule
Transcript Length (words) & \num{456}     & \num{521}     & \num{263} \\
Video Length (min)        & \num{4.51}    & \num{5.71}    & \num{2.12}   \\
Likes                     & \num{17086}   & \num{64145}   & \num{1585} \\
Views                     & \num{564111}  & \num{1824114} & \num{39320} \\
No. of Comments           & \num{1022}    & \num{6098}    & \num{64} \\
Subscribers (channels)    & \num{4137773} & \num{11562832}& \num{533000} \\
\bottomrule
\end{tabular}
\end{adjustbox}
\centering
\caption{Descriptive statistics of the YouTube dataset}
\label{tab:dataset-stats}
\end{table}

\begin{table*}[htbp]
\centering
\begin{tabular}{lp{11cm}}
\toprule
Theory & Definition \\
\midrule
Critical Discourse Analysis (CDA) & Links linguistic choices and discourse structures to power and ideology \\
Elaboration Likelihood Model (ELM) & Frames persuasion via a central (argument-focused) or peripheral (cue-driven) route \\
Classical Rhetoric  & Defines persuasion through ethos (credibility), pathos (emotion), and logos (logic) \\
Speech Act Theory & Treats language as action by classifying utterances by their intended function \\
Extended Parallel Process Model (EPPM) & Explains how fear appeals are mediated by perceived threat versus efficacy \\

\bottomrule
\end{tabular}
\caption{Theoretical foundations of the persuasive tactics taxonomy}
\label{tab:theory-defs}
\end{table*}

\subsection{LLM Annotation Framework}

To analyze persuasive strategies in gomutra discourse, we draw on five foundational theories of persuasion: Speech Act Theory, Critical Discourse Analysis (CDA), the Extended Parallel Process Model (EPPM), the Elaboration Likelihood Model (ELM), and Classical Rhetoric (Table~\ref{tab:theory-defs}). Together, these frameworks emphasize how communication functions as action (Speech Act Theory), how discourse is embedded in ideology and power (CDA), how the balance of fear and efficacy shapes responses to health messages (EPPM), how audiences process arguments through central or peripheral cues (ELM), and how recurring appeals to ethos, pathos, and logos structure persuasion (Classical Rhetoric).

Guided by these theories, we operationalize a 14-category taxonomy of strategic persuasive tactics (Table~\ref{tab:cue-defs}). The taxonomy includes: Appeal to Authority, Anecdotal Evidence, Appeal to Tradition/Religion, Appeal to Modernity, Scientific Jargon, Conspiracy Framing, Binary Framing/Othering, Call to Action, Denialism/Rebuttal, Moral Obligation, Celebrity Endorsement, Fear Appeal, Efficacy Appeal, and Social Proof. These categories extend beyond surface-level linguistic features to capture how speakers establish credibility, mobilize audiences, discredit opponents, and anchor claims in cultural or emotional appeals. By unifying cues across multiple theoretical traditions, the framework enables systematic comparison of persuasion strategies at scale.

We applied this taxonomy across six large language models representing different architectures and training regimes: GPT-4, GPT-4o, GPT-4.1, GPT-5, Gemini 2.5 Pro, and Mistral Medium 3 (Table~\ref{tab:models}). Each model was provided with a standardized prompt that included definitions of the 14 tactics and the full transcript of each video. Models were asked to annotate each transcript with all applicable cues inline. The resulting annotations form the basis for two key analyses: (1) a cross-model comparison of cue distributions and sensitivities, and (2) human validation to assess alignment, precision, and the limits of automated rhetorical analysis.

We release the annotated dataset, licensed under CC BY 4.0. Author and repository details are withheld for double-blind review.

\begin{table}[ht]
\centering
\begin{tabular}{lc}
\toprule
LLM & Mean Cue Density (per 100 words) \\
\midrule
GPT-4            & 0.066 \\
GPT-4o           & 0.062 \\
GPT-4.1          & 0.154 \\
GPT-5            & 0.374 \\
Gemini 2.5 Pro   & 0.251 \\
Mistral Medium 3 & 0.159 \\
\bottomrule
\end{tabular}
\caption{LLMs and corresponding cue densities}
\label{tab:models}
\end{table}

\begin{table*}[t]
\centering
\begin{tabular}{lp{11cm}}
\toprule
Cue & Definition \\
\midrule
Appeal to Authority & Invoking experts, institutions, or leaders to validate claims \\
Anecdotal Evidence & Using personal stories or testimonials as proof \\
Appeal to Tradition/Religion & Validating beliefs by cultural or religious heritage \\
Appeal to Modernity & Framing credibility via progressiveness or global uptake \\
Scientific Jargon & Technical or pseudo-technical terms to suggest credibility \\
Conspiracy Framing & Suggesting elites, corporations, or outsiders suppress truth \\
Binary Framing/Othering & In-group vs. out-group divisions \\
Call to Action & Directing the audience to act, share, or adopt behavior \\
Denialism/Rebuttal & Rejecting or minimizing opposing evidence \\
Moral Obligation & Framing an action as an ethical or moral duty \\
Celebrity Endorsement & Using fame or influence as validation \\
Fear Appeal & Emphasizing threat severity or vulnerability \\
Efficacy Appeal & Showing how an action can reduce/prevent/fix a threat \\
Social Proof & Suggesting consensus, popularity, or widespread adoption \\
\bottomrule
\end{tabular}
\caption{Persuasive tactics taxonomy with definitions}
\label{tab:cue-defs}
\end{table*}

\newpage
\subsection{Prompt for LLM Annotation}

We instructed the LLMs with the following full prompt:

\newtcolorbox{PromptBox}{
  enhanced, breakable,
  width=\columnwidth,
  colback=gray!2, colframe=gray!55,
  colbacktitle=gray!8, coltitle=black,
  boxrule=0.5pt, arc=2mm,
  left=8pt, right=8pt, top=8pt, bottom=8pt,
  before skip=8pt, after skip=10pt
}

\newenvironment{PromptText}{%
  \ttfamily\footnotesize\linespread{1.03}\selectfont
  \setlength{\parindent}{0pt}%
   \raggedright
   \justifying 
  \setlength{\parskip}{0pt}%
 
}{}

\newcommand{\Tag}[2]{%
  \par\noindent\hangindent=2.2em\hangafter=1 [#1]\ #2%
}

\begin{PromptBox}
\begin{PromptText}

\begin{flushleft}{
\textbf{SYSTEM INSTRUCTION:}}\end{flushleft}

You are an expert in rhetorical and critical discourse analysis. Your task is to annotate strategic persuasive tactics in a transcript. The output must be the entire transcript, unchanged and unabridged, with inline labels in square brackets appended at the end of relevant sentences or phrases. Do not summarize, shorten, or paraphrase.

\vspace{0.1cm}

\begin{flushleft}{\textbf{USER INSTRUCTION:}}\end{flushleft}

Read the transcript carefully. Insert the following strategic persuasion tags where they occur. Each label should appear immediately after the sentence/phrase that exemplifies it.

\vspace{0.1cm}
\begin{flushleft}{
\textbf{LABELS \& DEFINITIONS}}\end{flushleft}
\begin{enumerate}
\item 
\Tag{Appeal to Authority}{Invoking experts, institutions, or leaders to validate claims.}
\item 
\Tag{Anecdotal Evidence}{Using personal stories or testimonials as proof.}
\item
\Tag{Appeal to Tradition/Religion}{Framing beliefs as valid because of cultural or religious heritage.}
\item
\Tag{Appeal to Modernity}{Framing something as credible because it is progressive, global, or cutting-edge.}
\item
\Tag{Scientific Jargon}{Using technical or pseudo-technical terminology to create credibility.}
\item
\Tag{Conspiracy Framing}{Suggesting elites, corporations, or outsiders suppress the truth.}
\item
\Tag{Binary Framing/Othering}{Splitting the world into in-groups vs. out-groups.}
\item
\Tag{Call to Action}{Directing the audience to act, share, or adopt behavior.}
\item
\Tag{Denialism/Rebuttal}{Rejecting, dismissing, or minimizing opposing evidence.}
\item
\Tag{Moral Obligation}{Framing an action as an ethical or moral duty.}
\item
\Tag{Celebrity Endorsement}{Using fame, influence, or reputation as persuasive validation.}
\item
\Tag{Fear Appeal}{Emphasizing threat severity or personal vulnerability.}
\item
\Tag{Efficacy Appeal}{Showing how an action can reduce, prevent, or fix a threat.}
\item
\Tag{Social Proof}{Suggesting consensus, popularity, or widespread adoption.}

\end{enumerate}
\begin{flushleft}{
\textbf{INSTRUCTION:}
Return the entire transcript with only these tags appended inline.}\end{flushleft}


\end{PromptText}
\end{PromptBox}


\section{Human Evaluation}
\subsection{Annotation Procedure}
To validate the persuasive tactics identified by LLMs, we conducted a comprehensive human evaluation study. Four independent annotators evaluated LLM-generated tags across our dataset of 100 YouTube transcripts, which contained 2,471 total LLM-generated persuasive tactic annotations. \\
For each LLM-generated tag, annotators provided binary judgements (Agree/Disagree) indicating whether they concurred with the LLM's identification of the specific persuasion tactic. This binary assessment enabled direct measurement of human-LLM alignment in persuasive tactic recognition \cite{mathew2021hatexplain, cris2023annotators}. Here, we distinguish between two evaluation metrics: \textit{acceptance}
measures human validation of LLM-generated tags (serving as a Human-LLM precision proxy), while \textit{agreement} captures inter-annotator reliability among human evaluators.

\subsection{Inter-annotator reliability (human-human)}
\begin{figure}[H]
  \centering
  \includegraphics[width=\linewidth]{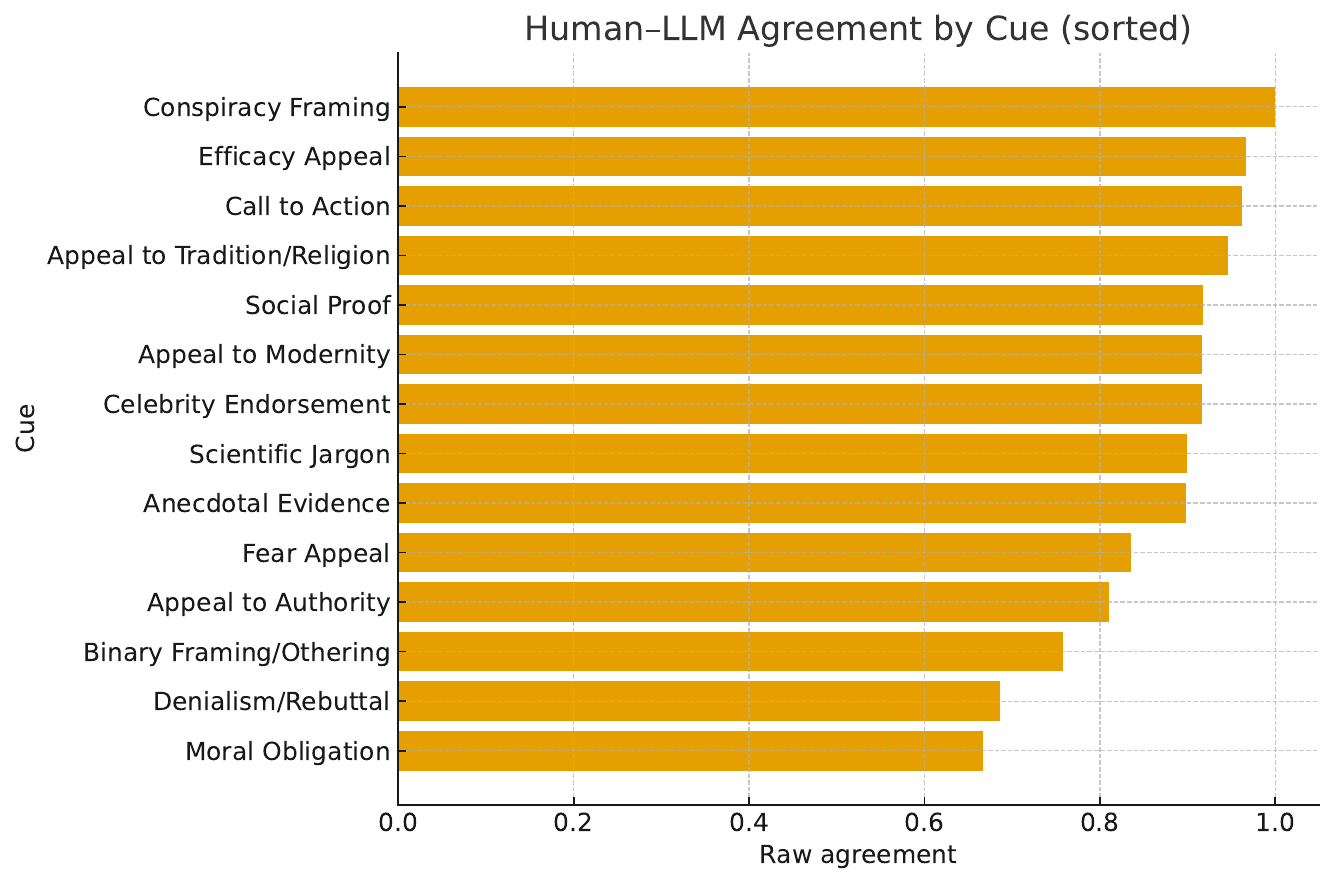}
  \caption{Human--human agreement by cue (inter-annotator reliability)}
  \label{fig:human-cue-bar}
\end{figure}

\textit{Overall reliability:}  The inter-annotator raw agreement was 0.901, across 2,471 decisions, suggesting a strong alignment. An examination of reliability across specific persuasive cues revealed significant variation. A subset of cues demonstrated high-reliability, including \emph{Conspiracy} (1.000), \emph{Efficacy} (0.967), \emph{Call to Action} (0.962), \emph{Tradition/Religion} (0.946), \emph{Social Proof} (0.918). Conversely, several cues exhibited lower levels of agreement: \emph{Denialism/Rebuttal} (0.686) and \emph{Moral Obligation} (0.667).  Similarly, \emph{Binary Framing/Othering} (0.758) achieved moderate reliability, suggesting that its operationalization requires refinement, particularly concerning the classification of pronoun usage, in-group/out-group references, and implicit audience construction (see Table \ref{tab:agreement_cue}
\& Figure \ref{fig:human-cue-bar}, Appendix). 

We also assessed inter-annotator reliability at the level of individual models' outputs and found high agreement across all models: \emph{Mistral Medium 3} (0.897, N=451)
\emph{Gemini 2.5 Pro}(0.899, N=757); \emph{GPT-4.1} (0.858, N=385); \emph{GPT-5} (0.854, N=1038); \emph{GPT-4o} (1.000, N=152) and \emph{GPT-4} (1.000, N = 162) (see Table \ref{tab:agreement_model}, Appendix).

\subsection{Acceptance as precision proxy (Human-LLM)}
Human acceptance of model-generated tags provides a practical proxy for precision in the absence of exhaustively annotated ground truth. In this study, agreement is defined as the proportion of model-assigned cues that were confirmed by at least one human annotator. While this does not address recall, it offers a robust measure of whether the cues a model surfaces are valid in the eyes of human evaluators.

\begin{figure}[ht]
  \centering
  \includegraphics[width=\linewidth]{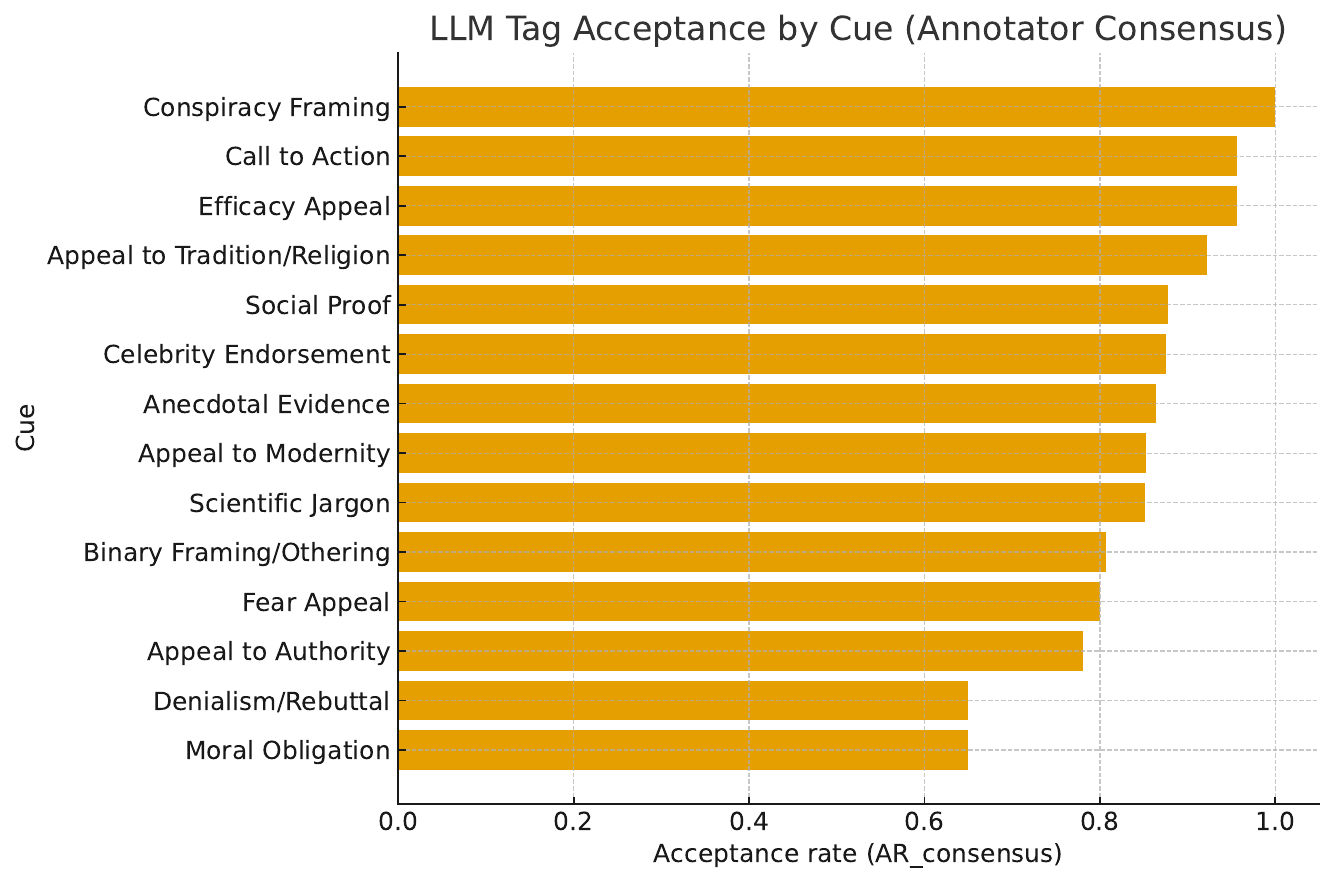}
  \caption{LLM tag acceptance by cue (AR\_consensus)}
  \label{fig:acc-cue-cons}
\end{figure}

\textit{Cue-level precision:} Agreement is highest for cues with clear and explicit lexical or semantic markers, such as \emph{Efficacy}, \emph{Call to Action}, and \emph{Tradition/Religion}. In contrast, more implicit or dialogic moves, including \emph{Denialism/Rebuttal} and \emph{Moral Obligation}, yielded lower agreement rates, reflecting both their conceptual ambiguity and the greater interpretive burden placed on annotators. This asymmetry highlights the importance of explicit operationalization for contested categories (see Figure \ref{fig:acc-cue-cons}, Table ~\ref{tab:acceptance_cue}).

\begin{table}[t]
\centering
\small
\begin{tabular}{lrrr}
\toprule
Cue & AR\_mean & AR\_consensus & N \\
\midrule
Efficacy Appeal & 0.968 & 0.956 & 627 \\
Appeal to Tradition/Religion & 0.955 & 0.923 & 431 \\
Appeal to Authority & 0.855 & 0.781 & 427 \\
Scientific Jargon & 0.888 & 0.852 & 374 \\
Call to Action & 0.970 & 0.957 & 365 \\
Anecdotal Evidence & 0.924 & 0.864 & 197 \\
Fear Appeal & 0.910 & 0.800 & 145 \\
Denialism/Rebuttal & 0.792 & 0.650 & 120 \\
Social Proof & 0.929 & 0.877 & 78 \\
Binary Framing/Othering & 0.800 & 0.806 & 45 \\
Appeal to Modernity & 0.920 & 0.853 & 44 \\
Conspiracy Framing & 1.000 & 1.000 & 37 \\
Celebrity Endorsement & 0.800 & 0.875 & 30 \\
Moral Obligation & 0.833 & 0.650 & 24 \\
\bottomrule
\end{tabular}
\caption{Acceptance rates (precision proxy) by cue}
\label{tab:acceptance_cue}
\end{table}

\textit{Model-level precision:} Differences also emerge across models. \emph{Mistral Medium 3} and \emph{Gemini 2.5 Pro} achieve the strongest average agreement rates (AR\_mean), suggesting high precision in their cue predictions. By contrast, \emph{GPT-4.1} and \emph{GPT-5} show slightly lower agreement, which appears to reflect a trade-off: these models surface a denser set of cues, thereby broadening coverage but at the cost of introducing more borderline cases (Table ~\ref{tab:acceptance_model}).

\begin{table}[htbp]
\centering
\small
\begin{tabular}{lrrr}
\toprule
Model & AR\_mean & AR\_consensus & N \\
\midrule
Mistral Medium 3 & 0.940 & 0.889 & 451 \\
Gemini 2.5 Pro   & 0.930 & 0.882 & 757 \\
GPT-4.1          & 0.917 & 0.844 & 385 \\
GPT-5            & 0.911 & 0.854 & 1038 \\
GPT-4o           & 0.901 & 0.901 & 152 \\
GPT-4            & 0.895 & 0.895 & 162 \\
\bottomrule
\end{tabular}
\caption{Acceptance rates (precision proxy) by model}
\label{tab:acceptance_model}
\end{table}

\textit{Evaluation scope:} It is important to stress that the validation conducted here is \emph{precision-oriented}. We do not report completeness-oriented metrics such as recall or F1, since annotators only evaluated cues proposed by the models rather than exhaustively annotating every possible cue in the transcripts. This design makes agreement a reliable upper-bound estimate of precision, but not a full measure of retrieval quality.


\section{Findings}
  


\subsection{Cue distribution by stance}
\begin{figure}[t]
  \centering
  \includegraphics[width=\linewidth]{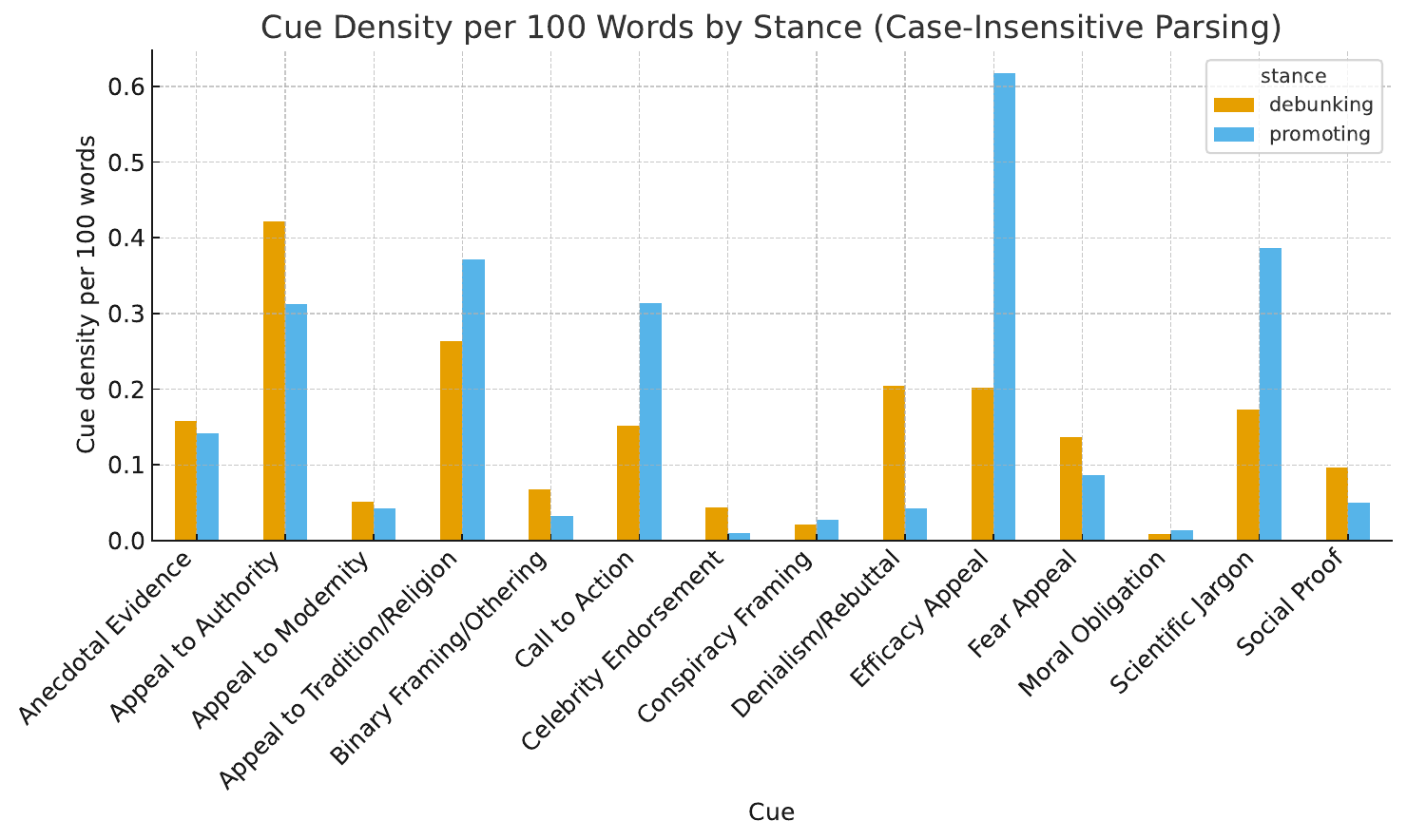}
  \caption{Cue density per 100 words by stance}
  \label{fig:cue-distribution}
\end{figure}

Analysis of cue distributions reveals clear rhetorical splits between promoting and debunking content. Promoting videos concentrate heavily on \emph{Efficacy Appeals}, \emph{Appeals to Tradition/Religion}, and \emph{Calls to Action}, while debunking videos are more likely to emphasize \emph{Appeals to Authority}, \emph{Denialism/Rebuttal}, and \emph{Scientific Jargon} (Figure~\ref{fig:cue-distribution}, Table~\ref{tab:stance_full_density}).

This asymmetry is statistically significant: promoters legitimate and mobilize their audiences through efficacy, tradition, and action (median $U = 988.0$, $p < 0.05$), while debunkers credential and refute through authority, rebuttal, and jargon (median $U = 993.5$, $p < 0.05$; see Appendix Table~\ref{tab:functional_asymmetry_tests}). 

The practical implication is that corrective messaging which relies solely on factual rebuttals may underperform unless it also addresses the efficacy and social proof frames that drive the appeal of promoting content. In short, the persuasive ecosystem is functionally asymmetric: promoters rely on mobilizing tactics, while debunkers rely on credentialing and refutation.

\begin{table*}[t]
\centering
\begin{tabular}{lrr}
\toprule
Cue & Debunking & Promoting \\
\midrule
Appeal to Authority & 0.443 & 0.331 \\
Anecdotal Evidence & 0.166 & 0.151 \\
Appeal to Tradition/Religion & 0.277 & 0.394 \\
Appeal to Modernity & 0.054 & 0.045 \\
Scientific Jargon & 0.181 & 0.409 \\
Conspiracy Framing & 0.022 & 0.028 \\
Binary Framing/Othering & 0.072 & 0.034 \\
Call to Action & 0.160 & 0.334 \\
Denialism/Rebuttal & 0.215 & 0.045 \\
Moral Obligation & 0.008 & 0.015 \\
Celebrity Endorsement & 0.046 & 0.011 \\
Fear Appeal & 0.144 & 0.091 \\
Efficacy Appeal & 0.211 & 0.655 \\
Social Proof & 0.102 & 0.053 \\
\bottomrule
\end{tabular}
\caption{Cue density per 100 words by stance (full 14-cue table; words computed after stripping inline tags)}
\label{tab:stance_full_density}
\end{table*}

\subsection{Model divergence (density profile)}


The models differ substantially in how densely they annotate cues, with important consequences for the kinds of signals surfaced. Higher-density models such as GPT-5 and Gemini 2.5 Pro are more likely to capture marginal cues (median $U = 7295.5$, $p < 0.001$; see Appendix Table~\ref{tab:marginal_cues_density_test}), whereas conservative models such as GPT-4 and GPT-4o tend to flag only the most salient signals (medians 8.0 vs.\ 2.0 per transcript). 

\begin{figure}[t]
  \centering
  \includegraphics[width=\linewidth]{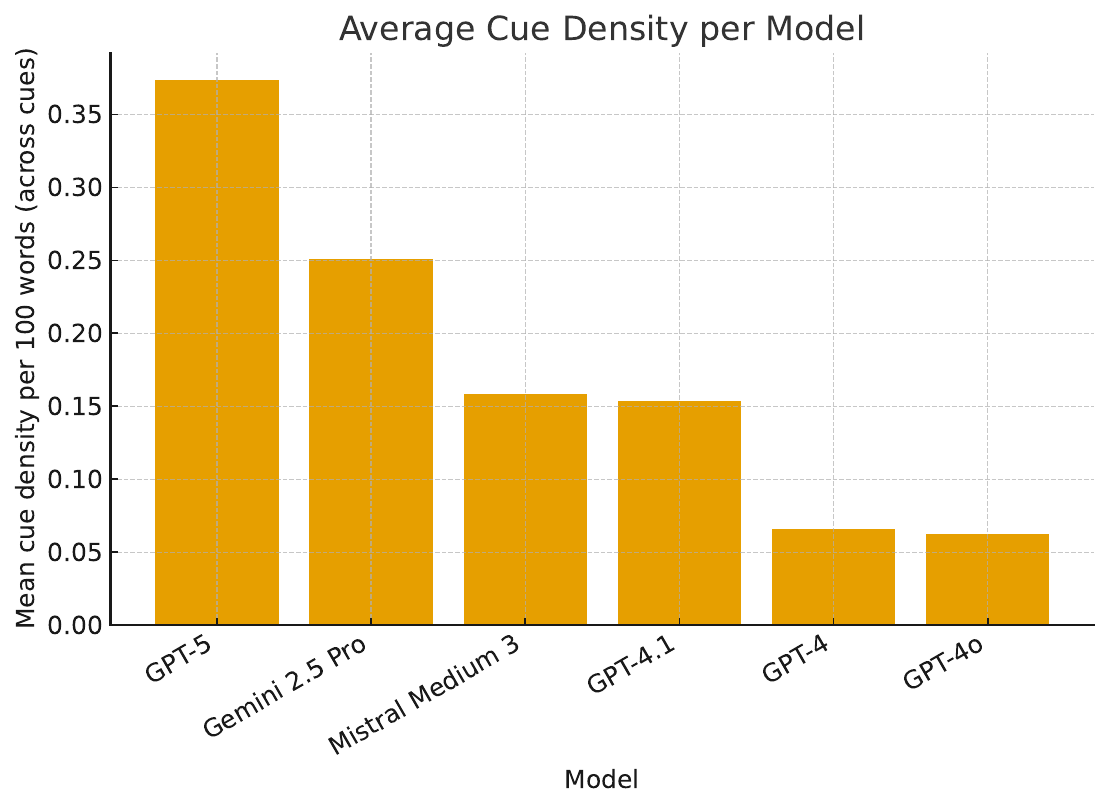}
  \caption{Average cue density per model (mean across cues)}
  \label{fig:model-comparison}
\end{figure}

This divergence is not only a function of overall tagging volume but also reflects differences in the detection of culturally specific cues, such as community-based appeals and tradition-linked references. Cue-specific sensitivity further illustrates these differences: \emph{Scientific Jargon} and \emph{Efficacy Appeals} show the largest between-model spread, underscoring that modeling choices strongly influence these categories (Figure~\ref{fig:model-comparison}, Table~\ref{tab:model_full_density}). 

For applied use, the trade-off is clear: density-forward models are preferable when comprehensive coverage and exploratory insight are desired, while conservative models are more suitable for benchmarking or low-supervision analysis where precision is paramount.


\begin{table*}[t]
\centering
\begin{tabular}{lrrrrrr}
\toprule
Cue & GPT-4 & GPT-4o & GPT-4.1 & GPT-5 & Gemini 2.5 Pro & Mistral Medium 3 \\
\midrule
Appeal to Authority & 0.166 & 0.216 & 0.502 & 0.606 & 0.310 & 0.327 \\
Anecdotal Evidence & 0.108 & 0.106 & 0.180 & 0.234 & 0.175 & 0.120 \\
Appeal to Tradition/Religion & 0.214 & 0.248 & 0.397 & 0.606 & 0.400 & 0.360 \\
Appeal to Modernity & 0.025 & 0.029 & 0.041 & 0.095 & 0.053 & 0.038 \\
Scientific Jargon & 0.092 & 0.101 & 0.338 & 0.991 & 0.402 & 0.265 \\
Conspiracy Framing & 0.016 & 0.005 & 0.027 & 0.045 & 0.046 & 0.022 \\
Binary Framing/Othering & 0.007 & 0.005 & 0.037 & 0.069 & 0.091 & 0.036 \\
Call to Action & 0.069 & 0.054 & 0.168 & 0.864 & 0.471 & 0.162 \\
Denialism/Rebuttal & 0.037 & 0.027 & 0.032 & 0.149 & 0.135 & 0.082 \\
Moral Obligation & 0.000 & 0.002 & 0.005 & 0.014 & 0.029 & 0.029 \\
Celebrity Endorsement & 0.007 & 0.005 & 0.012 & 0.033 & 0.024 & 0.024 \\
Fear Appeal & 0.009 & 0.000 & 0.012 & 0.232 & 0.279 & 0.062 \\
Efficacy Appeal & 0.147 & 0.074 & 0.372 & 1.148 & 0.977 & 0.647 \\
Social Proof & 0.025 & 0.002 & 0.032 & 0.144 & 0.117 & 0.047 \\
\bottomrule
\end{tabular}
\caption{Cue density per 100 words by model (aggregated across videos)}
\label{tab:model_full_density}
\end{table*}

\subsection{Agreement (precision proxy) patterns in human evaluation}

\begin{figure}[htbp]
  \centering
  \includegraphics[width=\linewidth]{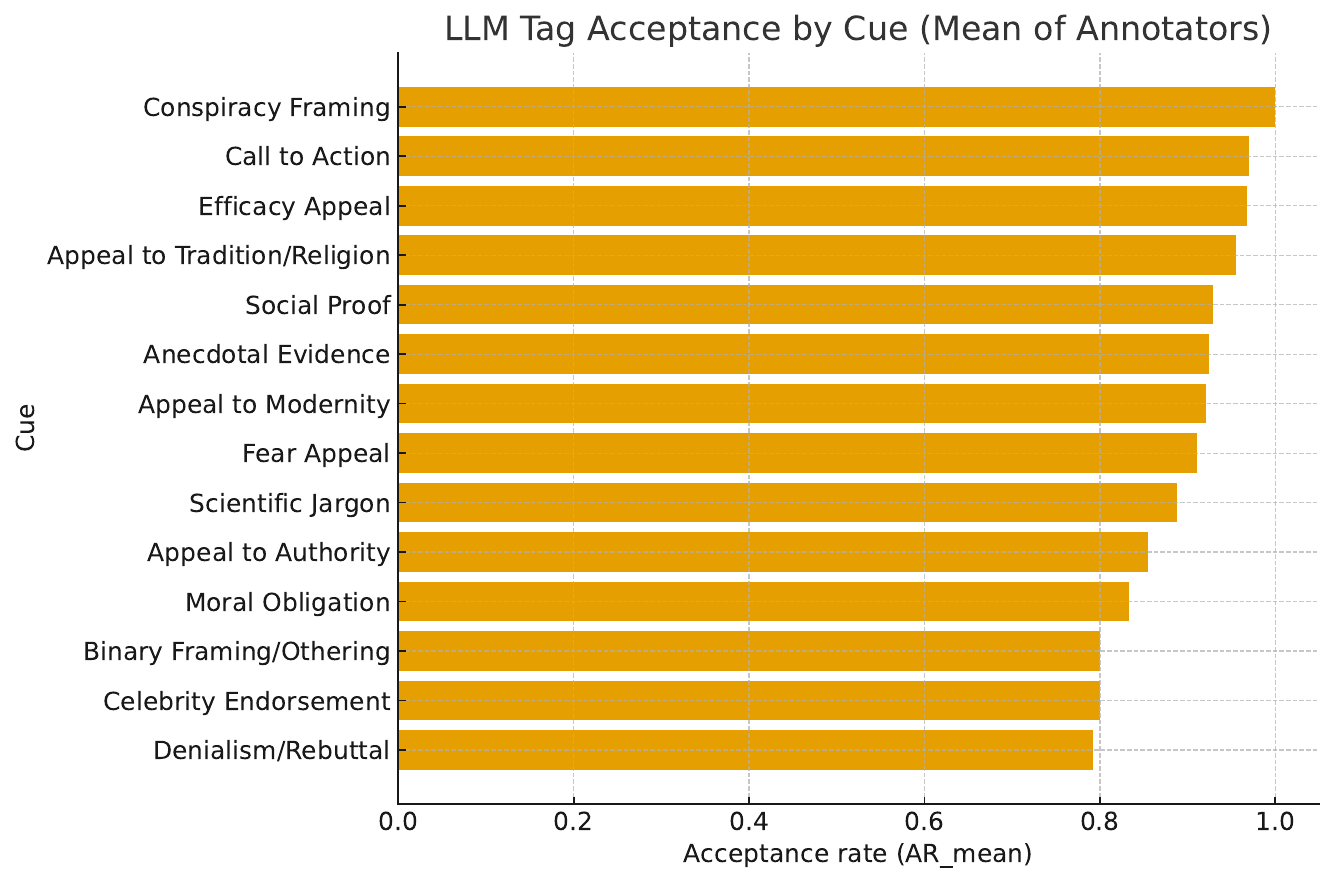}
  \caption{LLM tag acceptance by cue (AR\_mean)}
  \label{fig:acc-cue-mean}
\end{figure}

Human validation of model-generated tags provides a precision-oriented view of model performance. Cue-level analysis shows very high acceptance for \emph{Efficacy Appeals} (AR\_mean = 0.968), \emph{Calls to Action} (0.970), and \emph{Appeals to Tradition/Religion} (0.955), suggesting that LLMs are highly reliable for these surface-expressed, lexically explicit strategies (Figure~\ref{fig:acc-cue-mean}, Table~\ref{tab:acceptance_cue}).

By contrast, implicit or discourse-structured moves such as \emph{Denialism/Rebuttal} (AR\_mean = 0.792; AR\_consensus = 0.650) and \emph{Moral Obligation} (0.833; 0.650) yielded lower agreement, reflecting their greater interpretive burden and cultural specificity. 

At the model level, precision ordering places \emph{Mistral Medium 3} (AR\_mean = 0.940) and \emph{Gemini 2.5 Pro} (0.930) at the top, followed closely by \emph{GPT-4.1} (0.917) and \emph{GPT-5} (0.911), with GPT-4 and GPT-4o performing slightly lower but consistently on smaller evaluation subsets (Table~\ref{tab:acceptance_model}). Together, these results suggest that while LLMs can serve as effective first-pass annotators for explicit cues, human oversight remains essential for culturally loaded or implicit categories.

\section{Discussion}

Our findings show, across the LLMs, that the greatest strength of qualitative research as it relates to systematic analysis of large texts is in identifying surface-level, explicit signals. Semantically straightforward tactics such as implying efficacy through synonyms or easily identifiable intensifiers (e.g., adverbs) are almost always labeled correctly. However, when meaning is implied through more subtle tactics, LLMs do not perform as well. In other words, the motivation of the speaker may determine how effectively the LLM interprets meaning. On the side making claims about cow urine, assertions about benefits tend to be relatively direct, phrased in unambiguous language with distinct linguistic features that are straightforward to identify. By contrast, refutation or debunking is more complex, since it often mixes cues, sarcasm, or contrarian framing. Here, persuasion may be implied rather than directly stated, using a blend of direct language and innuendo that models do not consistently capture. Factual corrections that lack rhetorical counter-appeals are consequently underemphasized in LLM-based classification, even though those rhetorical appeals, often sarcastic, humorous, or offensive, may be precisely what makes a message resonate with audiences.

We also see that promoters and debunkers do not simply offer opposing facts, but that persuasion in their content is carried out through distinct rhetorical forms. Promoters lean more heavily on tactics that enable mobilization by highlighting efficacy, tradition, and action, while debunkers rely on credentials and refute using authority and jargon. The use of anecdote on the promoter side further strengthens rhetorical emphasis, allowing influencers and interlocutors to highlight cures through stories. Anecdotes are powerful because they are narratives, which naturally appeal to audiences. These findings suggest that interventions to strengthen information quality cannot rely on factuality alone, but must also contend with the tactical forms of affective information. This work also shows the importance of tone in outreach. In this sample, debunkers, who are themselves influencers and adept at playing the online “game”, tend to adopt a more authoritative tone. This stands in contrast to standard scientific debunking, where officials typically rely on bland facts and figures. Effective counter-misinformation may therefore require creativity in delivery, such as incorporating narrative elements, making value-based arguments, or presenting testimonials from individuals for whom remedies failed. As seen among successful influencer debunkers, they explain the logic of their position rather than relying on facts alone, underlining the importance of engaging audiences in reasoning rather than merely providing factual pushback.

The training of LLMs on culturally specific content is crucial for the kinds of interpretive qualitative analysis they can provide. Many persuasive claims are implied rather than directly stated, and often use culturally rooted references or turns of phrase. Training data that encodes such cultural cues is therefore necessary for deeper analysis. Given that most training data are in English, the interpretive value of current LLMs for qualitative research on non-Western contexts is undermined. In this case, cultural specificity lies not only in traditional conceptions and uses of cow urine but also in the political meanings it carries. For those selling curative ideas around cow urine, the value proposition is also based on the trust networks driven by the political leaning of the audience, and by extension, on the preeminence of nativist beliefs. Debunkers, in turn, frame their audiences as politically motivated rather than logically persuaded. Going beyond surface-level persuasive tactics and capturing culturally embedded cues highlights why human readers remain essential, or at minimum why culturally informed prompt design and training on diverse corpora are necessary. Consistently training LLMs on culturally varied datasets is not only possible but arguably essential for greater interpretive fidelity in qualitative work.

Finally, the human in the loop makes a substantial difference to the quality of findings. In this study, most researchers were familiar with the South Asian context, the meaning and use of cow urine, and the extent to which folk remedies are institutionalized in practice. Without human involvement in qualitative analysis, major gaps in understanding would remain. A pragmatic model for analysis involves: (i) dense models generating an initial round of results; (ii) automatic flagging of low-precision or low-reliability cues; (iii) human adjudication; (iv) prompt or taxonomy updates; and (v) re-running annotations. This retains scale while ensuring cultural sensitivity. The broader point is that while humans remain indispensable for interpretation, LLMs can significantly extend systematic qualitative research by providing fine-grained evidence at a scale unattainable through manual coding. LLMs also integrate theoretical constructs effectively: our choice to frame prompts with “You are an expert in rhetorical and critical discourse analysis” introduced perspectives often new to us, highlighting how qualitative studies are inevitably framed by the researcher’s theoretical comfort. Importantly, LLMs partnered with trained human researchers can encode and apply complex theoretical constructs (e.g., “appeal to tradition,” “moral obligation”) across large datasets, operationalizing theory into the pattern-matching tasks that models excel at. What LLMs enable is not only the scaling of qualitative research but also the exploration of different meanings or interpretations of misinformation through the lens the researcher chooses. This extends the ability of human researchers to critically examine their own annotation choices and interpretive frames. Ultimately, the human remains essential for interpretation, but the labor-intensive work of coding and aggregation is meaningfully strengthened by the machine as assistant.

\section{Conclusion}

This study advances a rhetorically grounded approach to analyzing health misinformation on YouTube by operationalizing persuasive tactics and scaling annotation with large language models (LLMs) validated by human coders. We find a consistent asymmetry in tactic bundles: promoting videos emphasize efficacy, tradition, and calls to action, while debunking videos rely on authority, rebuttals, and scientific jargon. Human validators confirmed most LLM-assigned tags for explicit cues (e.g., \emph{Efficacy}, \emph{Call to Action}, \emph{Tradition/Religion}), but showed lower acceptance for implicit or discourse-structured categories (e.g., \emph{Denialism/Rebuttal}, \emph{Moral Obligation}). Agreement is therefore best interpreted as a precision-oriented metric. 
In practice, LLMs are effective first-pass annotators for surface cues, with human oversight essential for culturally loaded or ambiguous tactics. Interventions should target rhetorical forms, not only factual claims, and model choice should balance precision against coverage. Limitations include reliance on a single platform and context, small-sample cues, and a focus on precision rather than recall. Future work will expand the corpus across languages and modalities, refine culturally aware prompts, and calibrate model confidence for efficient human-in-the-loop workflows. Together, these findings underscore both the promise and the boundaries of LLMs in supporting culturally grounded analysis of online health misinformation.


\bibliography{reference.bib}

\newpage
\clearpage
\pagebreak
\section{Appendix}

\subsection{Statistical Significance of Findings Per Model}
\textbf{Functional asymmetry in cues}
\begin{samepage}
\begin{table}[!htbp]
\centering
\resizebox{\linewidth}{!}{%
\begin{tabular}{lll}
\toprule
\textbf{Model} & \textbf{Promoter cues (Promoting $>$ Debunking)} & \textbf{Debunker cues (Debunking $>$ Promoting)} \\
\midrule
GPT-4o              & $U = 792.0$, $p = 0.385$ (n.s.)   & $U = 996.5$, $p = 0.010^{**}$ \\
GPT-4               & $U = 937.0$, $p = 0.052$ ($\sim$) & $U = 865.5$, $p = 0.140$ (n.s.) \\
GPT-4.1             & $U = 946.0$, $p = 0.047^{*}$      & $U = 980.5$, $p = 0.023^{*}$ \\
GPT-5               & $U = 1035.5$, $p = 0.007^{**}$    & $U = 850.5$, $p = 0.211$ (n.s.) \\
Gemini 2.5 Pro      & $U = 1038.0$, $p = 0.007^{**}$    & $U = 937.0$, $p = 0.056$ ($\sim$) \\
Mistral Medium 3    & $U = 952.5$, $p = 0.043^{*}$      & $U = 1047.5$, $p = 0.005^{**}$ \\
\midrule
\textbf{Summary}    & Significant in 4/6 models, borderline in 1 & Significant in 3/6 models \\
\bottomrule
\end{tabular}
}
\caption{Mann--Whitney U tests of cue distributions by stance (Promoter cues = Efficacy Appeal, Appeal to Tradition/Religion, Call to Action; Debunker cues = Appeal to Authority, Denialism/Rebuttal, Scientific Jargon).}
\label{tab:functional_asymmetry_tests}
\footnotesize Notes: Mann--Whitney U test, one-sided. $^{*}p<0.05$, $^{**}p<0.01$, n.s. = not significant, $\sim$ = borderline.
\end{table}
\end{samepage}


\textbf{Density Differences}
\begin{table}[H]
\centering

\resizebox{\linewidth}{!}{%
\begin{tabular}{lcc}
\toprule
\textbf{Model} & \textbf{Mean tags/tx (density)} & \textbf{Marginal cues (High $>$ Low)} \\
\midrule
GPT-4o              & 3.31  & Lower group (conservative) \\
GPT-4               & 3.64  & Lower group (conservative) \\
GPT-4.1             & 8.40  & Lower group (conservative) \\
GPT-5               & 19.90 & Higher group (dense) \\
Gemini 2.5 Pro      & 13.44 & Higher group (dense) \\
Mistral Medium 3    & 8.92  & Higher group (dense) \\
\midrule
\textbf{Test (High vs.\ Low)} & Median split = 8.66 & $U = 7295.5$, $p = 1.14\times 10^{-9}$ \\
\textbf{Per-transcript counts} & High = 8.0 (median), 11.2 (mean) & Low = 2.0 (median), 3.3 (mean) \\
\bottomrule
\end{tabular}
}
\caption{Mann--Whitney U tests of marginal-cue counts by model density (Promoting vs.\ Debunking). Marginal cues = Binary Framing/Othering, Social Proof, Anecdotal Evidence, Call to Action.}
\label{tab:marginal_cues_density_test}
{\footnotesize Notes: “Density” = mean total tags per transcript. Marginal cues are the four least frequent globally. Mann--Whitney U compares transcript-level sums between high- and low-density model groups.}
\end{table}

\subsection{Human Annotation Details (human-human)}

\begin{table}[H]
\centering
\small
\begin{tabular}{lrrr}
\toprule
Cue & raw\_agreement & N \\
\midrule
Efficacy Appeal & 0.967 & 627 \\
Appeal to Tradition/Religion & 0.946 & 431 \\
Appeal to Authority & 0.810 & 427 \\
Scientific Jargon  & 0.899 & 374 \\
Call to Action & 0.962 & 365 \\
Anecdotal Evidence & 0.897 & 197 \\
Fear Appeal & 0.834 & 145 \\
Denialism/Rebuttal & 0.686 & 120 \\
Social Proof & 0.918 & 78 \\
Binary Framing/Othering & 0.758 & 45 \\
Appeal to Modernity & 0.916 & 44 \\
Conspiracy Framing & 1.000 & 37 \\
Celebrity Endorsement & 0.916 & 30 \\
Moral Obligation & 0.667 & 24 \\
\bottomrule
\end{tabular}
\caption{Agreement rates (human-human)by cue }
\label{tab:agreement_cue}
\end{table}

\begin{table}[H]
\centering
\small
\begin{tabular}{lrrr}
\toprule
Model & raw\_agreement & N \\
\midrule
Mistral Medium 3 & 0.897 &  451\\
Gemini 2.5 Pro   & 0.899 &  757\\
GPT-4.1          & 0.858 &  385\\
GPT-5            & 0.854 &  1038\\
GPT-4o           & 0.901 &  152\\
GPT-4            & 1.000 &  162\\
\bottomrule
\end{tabular}
\caption{Raw agreement rates (human-human) by model}
\label{tab:agreement_model}
\end{table}



\clearpage

\end{document}